\documentclass[sn-apa]{sn-jnl}

\usepackage{graphicx}%
\usepackage{multirow}%
\usepackage{amsmath,amssymb,amsfonts}%
\usepackage{amsthm}%
\usepackage{mathrsfs}%
\usepackage[title]{appendix}%
\usepackage{xcolor}%
\usepackage{textcomp}%
\usepackage{manyfoot}%
\usepackage{booktabs}%
\usepackage{algorithm}%
\usepackage{algorithmicx}%
\usepackage{algpseudocode}%
\usepackage{listings}%
\usepackage{tikz}%
\usepackage{float}%
\usetikzlibrary{arrows.meta,positioning}%
\usetikzlibrary{calc}%
\usepackage{afterpage}
\usepackage{pdflscape}
\usepackage{afterpage}
\usepackage{booktabs}
\usepackage{siunitx}
\begin{document}

\title[AI Generated Corpora]{AI Brown and AI Koditex: LLM-Generated Corpora Comparable to Traditional Corpora of English and Czech Texts}


\author*[1]{\fnm{Jiří} \sur{Milička}}\email{jiri@milicka.cz, ORCID: 0000-0001-8605-1199}

\author[1]{\fnm{Anna} \sur{Marklová}}\email{anna.marklova@ff.cuni.cz, ORCID: 0000-0003-3392-1028}

\author[1]{\fnm{Václav} \sur{Cvrček}}\email{vaclav.cvrcek@ff.cuni.cz, ORCID: 0000-0003-3977-2393}

\affil*[1]{\orgdiv{Department of Linguistics}, \orgname{Faculty of Arts, Charles University}, \orgaddress{\street{Panská}, \city{Prague}, \postcode{110 00}, \state{}, \country{Czech Republic}}}


\abstract{This article presents two corpora of English and Czech texts generated with large language models (LLMs). The motivation is to create a resource for comparing human-written texts with LLM-generated text linguistically. Emphasis was placed on ensuring these resources are multi-genre and rich in terms of topics, authors, and text types, while maintaining comparability with existing human-created corpora. These generated corpora replicate reference human corpora: BE21 by Paul Baker, which is a modern version of the original Brown Corpus, and Koditex corpus that also follows the Brown Corpus tradition but in Czech. The new corpora were generated using models from OpenAI, Anthropic, Alphabet, Meta, and DeepSeek, ranging from GPT-3 (davinci-002) to GPT-4.5, and are tagged according to the Universal Dependencies standard (i.e. they are tokenized, lemmatized, and morphologically and syntactically annotated). The subcorpus size varies according to the model used (the English part contains on average 864k tokens per model, 27M tokens altogether, the Czech partcontains on average 768k tokens per model, 21.5M tokens altogether). The corpora are freely available for download under the CC BY 4.0 license (the annotated data are under CC BY-NC-SA 4.0 licence) and are also accessible through the search interface of the Czech National Corpus.}


\maketitle

\section{Introduction}\label{sec1}
Current linguistic studies of texts generated by large language models tend to be experimental in nature, often adapting classic psycholinguistic experiments to this new domain (\cite{zhao2025large,sohail2025using}). Corpus linguistics has received less attention in this area, despite the clear potential for applying quantitative linguistic and corpus-based methodologies to LLM-generated texts.

At present, it is extremely difficult to find suitable data for standard corpus linguistic analysis: ideally, researchers need LLM-generated corpora that are directly comparable to human-generated reference corpora, allowing for meaningful comparisons between human and machine text production and between models.

Our newly created AI-Brown corpus addresses this need by building on BE21, a corpus from the Brown family compiled by Paul Baker (\cite{baker2023year}). BE21 represents a modern implementation of the original Brown Corpus methodology, maintaining similar composition and annotation standards while using contemporary texts.

We selected BE21 for several key reasons. First, it contains excerpts from a wide variety of text genres. Second, as a pre-2021 corpus, which means it is likely purely human-authored. Third, unlike the original Brown Corpus (\cite{francis1979brown}), BE21 contains modern English that closely matches the language found in contemporary LLM training datasets.

To create AI-Brown, we split each text chunk from BE21 into two parts: the first 500 words serves as a generation prompt, while the rest provides human-authored text for comparison.

Since most of the world's population are non-native English speakers, yet most LLMs were trained predominantly on English texts, examining non-English generation capabilities is crucial. Czech provides a good test-case: medium-sized language (approximately 10 million speakers) with well-developed language resources, yet it has (very probably) small representation in training data of all major LLMs. For prompting, we used the Koditex corpus from the Czech National Corpus, which is  comparable to BE21 in that it follows Brown Corpus design principles (\cite{zasina2019koditex}, \url{https://wiki.korpus.cz/doku.php/en:cnk:koditex}).

We generated texts using 13 frontier models from OpenAI, Anthropic, Meta, Alphabet, and DeepSeek across various parameter settings. We plan to expand the corpus to include new models as they emerge. Given the large number of available models, we focused on frontier models (those considered state-of-the-art at release, such as GPT-4) and widely-used models (like GPT-3.5-turbo, which was never the largest or strongest model, but served as ChatGPT's default for free users and was many people's first exposure to LLMs). We prioritized closed models accessible only through APIs, as these often become unavailable over time (as occurred with davinci-001). Open-weight models can be regenerated later with lower environmental impact as computational efficiency improves.

The resulting corpus serves dual purposes: comparing human versus machine text generation, and enabling cross-model comparisons.

The dataset includes raw JSON data containing generated texts, logprobs, alternative tokens, and complete API communications. The raw data are further processed: all texts are cleaned and annotated using Universal Dependencies standards, available in both plain text and CoNLL-U formats. 

Where possible, we used fixed random seeds to ensure reproducibility. While this was only available through OpenAI's API (other providers don't support this feature), we recorded all hyperparameter values to maximize reproducibility across platforms.

The dataset is released under open licenses: CC BY 4.0 for texts and raw data, accompanied by all generation scripts, CC BY-NC-SA 4.0 for annotated materials.

This represents the first corpus of its kind, with continuous development beginning in spring 2024. As such, it serves as an archive for models that disappeared from APIs or undergo modifications over time.

\section{Related Works}

Naturally, there exists a vast quantity of datasets generated during language model evaluations, dating back to before these models were termed ``large" (the first language model output can be found in Shannon's seminal work \cite{shannon1948mathematical}).

However, until approximately the GPT-2 era, these models were primarily worth examining as components of larger systems or tools, such as translators or other applications, because standalone language models could not produce longer coherent texts. Without user intervention, the limited context window caused texts to drift thematically and stylistically from the original prompt in unpredictable directions.

From this earlier period, various evaluation datasets exist for machine-translated texts and comparative corpora of machine-translated content for translation studies purposes, such as \cite{lapshinova2013vartra}.

Since 2020, when GPT-3 emerged as the first LLM to be commercially deployed as a language model in today's sense, numerous informal databases of generated texts have appeared (for example, Infinite Backrooms, \url{https://dreams-of-an-electric-mind.webflow.io/}). However, these are not curated or collected as proper corpora (i.e., they lack systematic methodology, clear collection and preservation procedures, and linguistic annotation). 

Evaluation-based datasets have also proliferated; thousands of outputs from benchmarks like MMLU \cite{hendrycks2020measuring} have likely been generated over the past five years. However, these datasets are not ideal for linguistic research: first, companies often train directly on benchmarks to achieve better scores \cite{golchin2023time}, which makes them not representative to normal outpusts, second, benchmarks typically contain easily evaluable (ideally automatically assessable) task sets, while linguistic research requires greater textual variation without strict task constraints.

Various databases have emerged for training human/AI classifiers or ad-hoc corpora for contrastive linguistic analysis (for example, a solid open corpus can be found in the data accompanying \cite{munoz2024contrasting}, though this corpus is small and single-genre). This is also the case with our corpus, which started as an ad-hoc corpus for a small stylometric study presented at ICAME45 (\cite{Milicka2024RegisterVariation}), but then evolved into a much larger project. We can assume that companies operating commercial human/AI classifiers possess extensive corpora required for training, but understandably, these are not made publicly available.

A promising attempt at creating a generated corpus in the corpus linguistics sense is the \emph{Fakecorp} corpus by Slovak Academy of Sciences. Unfortunately, the brief description at \url{https://www.juls.savba.sk/fakecorp.html} does not clarify the exact collection methodology nor identify its author (only a disclaimer that the Slovak National Corpus is \emph{not} its author). The corpus was generated using gpt-4o-mini-2024-07-18 (sampling parameters not specified). Based on the search interface, the Slovak corpus contains approximately 26M tokens.

\section{Methods}\label{sec:methods}

\subsection{Generating Procedure}

The core idea behind the corpus is to probe many \emph{diverse} regions of each model’s latent space. Rather than using short instruction prompts, we seed models with longer source texts and ask for continuation. A natural source of long, heterogeneous stimuli is some of the Brown family of reference corpora, which sample many registers with matched designs. From this family we selected the British English 2021 Corpus (BE21)  \citep{baker2023year}, which is searchable via CQPweb at \url{https://cqpweb.lancs.ac.uk/be2021/}
\citep{hardie2012cqpweb}. For Czech we used \textit{Koditex} \citep{zasina2019koditex}, a Brown-inspired corpus whose metadata scheme is tailored to Czech genres/registers; the corpus is accessible through the KonText interface at \url{https://www.korpus.cz/kontext/query?corpname=koditex}
\citep{machalek2020kontext}.

\begin{figure}[ht]
    \centering
\centering
\includegraphics[width=0.99\linewidth]{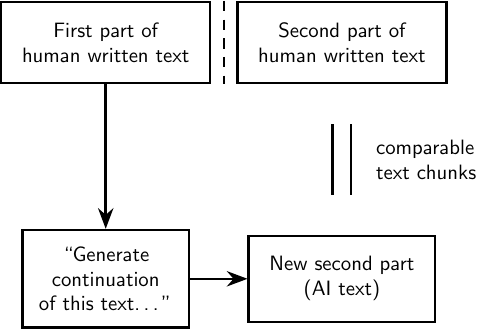}
    \caption{Schema of generating the corpus.}
    \label{fig:corpus_generation_schema}
\end{figure}

To obtain a directly comparable held-out reference, each source \emph{text chunk} was split into two parts: the first one supplied to the model and the second one reserved as the reference part for evaluation and qualitative comparison. The procedure is summarized schematically in Figure~\ref{fig:corpus_generation_schema}.

\begin{itemize}
\item \textbf{Prompt portion}: The first 500 words (including punctuation) served as generation prompts
\item \textbf{Reference portion}: The remaining text (approximately 1,500 words for English, variable length for Czech) provided human-authored comparison material
\end{itemize}

This segmentation strategy ensures that models receive sufficient context for generation while maintaining substantial reference text for comparative analysis. Also, the context of 500 words left sufficient space in the context window even for older models (davinci-002 has maximum context of 2049 tokens, while 500 English words takes about 670 tokens)

The Czech Koditex corpus initially contained more text samples than BE21, in some cases more than one samples from one text. To maintain comparability, we selected written texts only, one sample per source text to avoid overrepresentation, also all translated texts were omited since we wantd to avoid machine translated and machine-assisted translated texts in the sample.

The final Czech dataset contains 676 text samples, while the English dataset contains 500 samples, reflecting the original BE21 structure.

\subsection{Models and modes}

The corpus aims to cover models that were, at their time, considered break-through or state-of-the-art systems. Consequently, our sample primarily includes proprietary models accessible only via API. We make two exceptions for open-access models: \textit{Llama-3.1 405B} and \textit{DeepSeek-V3}. Both generated substantial public interest upon release, and including them allows us to study open models that were widely discussed as competitive with frontier systems. \citep{meta_llama31_blog,meta_llama31_hf,deepseek_v3_arxiv}

Our general policy is to \emph{not} routinely sample open-access models when we can archive them in full; open weights can be preserved for later re-generation of corpora once energy costs (and environmental impacts) decline. By contrast, models gated behind APIs tend to be modified or discontinued and must be sampled contemporaneously—despite higher financial and energy costs, if one wishes to preserve their behavior. For example, we missed the oportunity to capture the original GPT-3 model (\texttt{davinci-001}) despite its historical importance as the first widely accessible LLM to spark mainstream interest in 2020 \citep{brown2020gpt3}. The model is probably lost forever.

\subsubsection{OpenAI}
Originally a non-profit organization that has gradually transformed into a for-profit entity. The first to fully embrace the idea that mere scaling transformer architecture \citep{vaswani2017attention} could achieve intelligent behavior, which proved relatively capable with GPT-2 (February 2019) \cite{solaiman2019gpt2release}. Since then, they have remained at the forefront of LLM and adjacent technology development. Despite the name, the models are closed behind an API and details about their training (architecture details and training data) are not available.

OpenAI uses reinforcement learning from human feedback (RLHF) for fine-tuning, aiming to achieve a default ``helpful assistant'' persona while refusing to assist with harmful tasks or things considered unethical by the corporation \citep{openai_lp_2019,radford2019gpt2,openai_gpt4_techreport,ouyang2022rlhf}.

\paragraph{Davinci-002} (\texttt{davinci-002}, released 2023-08-22, knowledge cutoff date June 2021, discontinuted early 2025). Not exactly the original GPT-3 (which is davinci-001), but it has all its characteristics --- it works in completion mode, is not instruction-tuned, and can be considered a base model.

\paragraph{GPT 3.5 turbo} (\texttt{gpt-3.5-turbo-16k-0613}, released 2023-06-13, knowledge cutoff  September 2021)
For many users this was the first LLM they ever used, as it powered the free tier of ChatGPT for a long period. We used it both for standard generation and to reproduce the original ChatGPT system-prompted style. \citep{openai_0613,openai_chatgpt}

\paragraph{GPT-3.5 turbo in completion mode} (\texttt{gpt-3.5-turbo-instruct}, released 2023-08-22, knowledge cutoff unknown) As the full name suggests, this is not a base model but an instruction-tuned model that was available in completion mode (i.e., it autonomously completed text without a system prompt).

\paragraph{GPT-4} (\texttt{gpt-4-0613}; released 2023-06-13; knowledge cutoff: Sep 2021) GPT-4 replaced GPT-3.5 in ChatGPT’s paid tier and saw limited competition for roughly a year from its initial March 2023 launch. \citep{openai_gpt4_techreport,openai_0613}.

\paragraph{GPT 4 Turbo} (e.g., \texttt{gpt-4-1106-preview}, later \texttt{gpt-4-turbo-2024-04-09}; announced 2023-11-06; knowledge cutoff: Apr 2023).
A smaller GPT-4 variant. Comparing GPT-4-Turbo to GPT-4 reveals trade-offs introduced by optimization, or what was sacrificed to make the model run faster. \citep{openai_devday}

\paragraph{GPT-4o} (\texttt{gpt-4o}; announced 2024-05-13; knowledge cutoff later extended to Jun 2024) The first model to include a visual component. It was included in the corpus, because it was the default offering in ChatGPT for a long time. ; OpenAI iterated it frequently. \citep{openai_release_notes}

\paragraph{GPT-4.5} (\texttt{gpt-4.5-preview-2025-02-27}; released 2025-02-27; knowledge cutoff: not officially disclosed; third-party analyses suggest Oct 2023).
The newest and most expensive OpenAI chat model during spring 2025; we captured it before preview deprecation in mid-2025. \citep{openai_gpt45_card,openai_gpt45_blog,openai_gpt45_depr}

GPT-4.1 and GPT-5 will be included in the subsequent versions of the corpus.

\subsubsection{Anthropic}
Anthropic started with language models in 2021, but entered the mainstream consumer market only with Claude 3 Opus, which was in many ways better than the then-winner GPT-4. This model was particularly praised for its friendly and balanced persona, likely achieved through a different approach to additional training, as Anthropic employed Constitutional AI with RLAIF, using model-generated feedback guided by explicit principles.\citep{bai2022_constitutional,anthropic_claude3,anthropic_claude35_sonnet}.

In later years, Anthropic has focused especially on training models as tools for software development, while still maintaining high capabilities in other aspects.

\paragraph{Claude 3 Opus} (\texttt{claude-3-opus-20240229}; released 2024-03-04; knowledge cutoff: Aug 2023)
Probably Anthropic's most famous model. In its time, it was large and expensive, but contested GPT-4 not only in several benchmarks but also in popularity among power-users. \citep{anthropic_claude3}

\paragraph{Claude 3 Haiku} (\texttt{claude-3-haiku-20240307}; released 2024-03-07; knowledge cutoff: Aug 2023).
Anthropic’s smallest and cheapest Claude 3 model. \citep{anthropic_claude3}

\paragraph{Claude 3.5 Sonnet} (\texttt{claude-3-5-sonnet-20240620}; released 2024-06-20 knowledge cutoff April 2024)
Continuation of Claude 3. Claude 3.5 in the Opus version was never released, followed instead by Claude 4 Opus, which we don't yet have in the corpus and will be included in the next version.

\subsubsection{DeepSeek}
Chinese labs became increasingly competitive from the early 2020s, initially constrained by compute access. A turning point was \emph{DeepSeek-V3} (DeepSeek AI), widely discussed as SoTA among open models and available for local use. Its outputs exhibit political and ethical alignment choices that differ subtly from Western providers, making the corpus valuable for cross-cultural alignment analysis. \citep{deepseek_v3_arxiv}

\paragraph{DeepSeek-v3} (\texttt{deepseek-chat}; released 2024-12; knowledge cutoff: commonly reported Jul 2024 based on the exposed system prompt)
Reasoning variants (e.g., \emph{DeepSeek-R1}, released 2025-01) produce explicit chain-of-thought traces, but we did not use them here. Instead, we employed the standard chat-tuned (non-reasoning) variant for corpus generation \citep{deepseek_v3_arxiv,deepseek_r1_2025}

\subsubsection{Alphabet (Google)}

Foundational LLM work emerged from Google/Alphabet research (e.g., word2vec \citep{mikolov2013word2vec} and the Transformer \citep{vaswani2017attention}); the company focused first on productized NLP (notably neural machine translation) without publicly releasing general-purpose chat models. Under competitive pressure, Google introduced \emph{Bard/LaMDA} and later the \emph{Gemini} series. \citep{wu2016gnmt,thoppilan2022lamda}

\paragraph{Gemini 1.5 Pro 002} (\texttt{gemini-1.5-pro-002}; released 2024-09-24; knowledge cutoff not specified).
Google's multimodal model. The \texttt{-002} update delivered improved long-context performance and served as a widely used production model through 2024–2025 (later marked ``legacy'' as 2.0/2.5 arrived). \citep{google_vertex_versions}

\paragraph{Gemini 2.0 Pro (experimental)} (\texttt{gemini-2.0-pro-exp-02-05}; released 2025-02-05; knowledge cutoff not specified).
Experimental but successfull successor of Gemini 1.5 introduced alongside Flash variants. \citep{google_devblog_2025feb}

\paragraph{Gemini 2.0 Flash (experimental)} (\texttt{gemini-2.0-flash-exp}; announced 2024-12-11; knowledge cutoff not specified).
Smaller version of the Gemini 2.0 Pro model \citep{google_devblog_2024dec}. Along with GPT-4-turbo and Claude 3 Haiku, it can serve for research into what changes occur during model shrinking and optimization, which is not so much technical research as societal, as companies can influence which capabilities are preserved and which are sacrificed during fine-tuning.

\subsubsection{Meta (Facebook)}
Meta focuses on ``social chatbots" and companions and using LLMs for curating Facebook timelines, but it has a history of experimental interfaces for research \citep{taylor2022galactica}. Some of its models are provided opne-weights under the Llama Community License for free.

\paragraph{Llama 3.1 405B (base, 16-bit quantization)} (\texttt{Meta-Llama-3.1-405B}; released 2024-07-23; knowledge cutoff: Dec 2023).
This model was included in the corpus as it was the first freely available model of this scale. Its importance lies mainly in being the largest base model available to date.

\paragraph{Llama 3.1 405B Instruct (16-bit)} (\texttt{Meta-Llama-3.1-405B-Instruct}; released 2024-07-23; knowledge cutoff: Dec 2023).
Instruction-tuned version of the previous model. Serves mainly for comparing texts created by the base model, i.e., how much influence instruction tuning has.

\subsection{Prompts and sampling temperatures}

For each model, we generated two versions of each corpus: one using sampling temperature $T=0$ (deterministic generation) and one using $T=1$ (the model’s original probability distribution). For GPT-3.5-Turbo, we additionally tested $T=0.5$.

For base models and models operating in completion mode (davinci-002, GPT-3.5-Turbo, Meta-Llama-3.1-405B), we supplied only the initial portion of each source text as input, allowing the models to behave as traditional language models performing next-token prediction.

For instruction-tuned models, we employed minimal system prompts requesting a long continuation of the given text. Without such prompts, the models’ default \emph{helpful assistant} persona tended to analyze, summarize, or answer questions about the source text rather than continue it.

For English generation (AI-Brown), we used the following system prompt:
\emph{Please continue the text in the same manner and style, ensuring it contains at least five thousand words. The text does not need to be factually correct, but please make sure it fits stylistically.}

Language-specific challenges arose for Czech generation. Some models refused to comply when given Czech system prompts, necessitating the use of the following English system prompt. Despite explicit instructions to generate Czech text, several models sometimes produced English or mixed-language output:
\emph{Please continue the Czech text in the same language, manner and style, ensuring it contains at least five thousand words. The text does not need to be factually correct, but please make sure it fits stylistically.}

To ensure reproducibility, we used random seed 42 for all OpenAI API calls (see \cite{openai_cookbook_seed} for details about reproducibility in higher temperatures). Other providers did not offer comparable deterministic options. For Llama, generations used 16-bit floating-point (fp16) quantization, the highest quality available in our setup.

One subcorpus in both English and Czech corpora was designed to mimic the default ChatGPT persona. The corresponding system prompt mirrors the leaked ChatGPT prompt (\cite{chatgpt_prompt}; unofficial but independently corroborated by \cite{chatgpt_prompt,chatgpt_promptRohit,chatgpt_promptDobos}). Because this authentic prompt lacks explicit continuation instructions, we appended a brief continuation directive after each text chunk. We tested placing this directive both before and after the seed text; when placed before, it was ineffective, producing roughly half the intended corpus size and a high rate of refusals, so we appended it after the prompt text chunk.


\subsection{Postprocessing}
\label{sec:postprocessing}
API responses were archived in full (JSON), including token-level logprobs \citep{openai_cookbook_logprobs} and alternative candidates where a provider exposed them, to enable subsequent analyses of uncertainty and calibration. In addition, we materialized plain-text renditions that are convenient for downstream corpus work. These plain texts were then cleaned and linguistically annotated and exported in two corpus-friendly formats: \emph{CoNLL-U} \citep{ud_conllu_format,demarneffe2021ud} and a \emph{verticalized} representation compatible with widely used corpus engines (\emph{Manatee}/\emph{CWB}) and interfaces such as \emph{KonText}/Sketch Engine. \citep{rychly2004manatee,rychly2007manatee,evert2011cwb,machalek2020kontext,kilgarriff2014sketch}

\subsubsection{Cleaning}

Instruction-tuned chat models frequently prefixed continuations with meta-preambles (e.g., \emph{``Certainly! Here is a continuation in a similar style:''}). They also occasionally refused to continue—for instance, with content-policy–motivated apologies and redirections. In the canonical pipeline, such meta-commentary was removed. If the residual continuation after removal fell below a minimal length threshold (defined ex ante for each language), the sample was excluded from the \emph{clean} partition but preserved in the raw archive.

In cases where a model substituted the requested continuation with a different, ostensibly ''safer'' story or exposition (rather than refusing outright), we retained the content. Our rationale is that these substitutions reflect the model’s post-training alignment rather than transport-layer artifacts and are therefore of interest to downstream users.

Cleaning necessarily trades off between competing research uses. For example, formulaic meta-preambles are undesirable for keyword extraction or stylistic modeling, but valuable for studying model persona and alignment behaviors. Therefore, we distribute both minimally processed \emph{raw} text and JSON, and a \emph{clean} partition with light, documented normalizations. Model-specific formatting quirks (e.g., extra blank lines produced by \texttt{gpt-3.5-turbo} at $T=1$) were preserved in the \emph{clean} partition to maintain behavioral fidelity; users who prefer stronger normalization can build on our provided scripts and/or revert to the raw data.

\subsubsection{Annotation}

We annotate at multiple linguistic layers (tokenization, sentence segmentation, UPOS/XPOS tags, lemmas, morphological features, and dependency trees) within the \emph{Universal Dependencies} (UD) framework \citep{demarneffe2021ud,nivre2017ud}. Processing is performed with \emph{UDPipe} (english-ewt-ud-2.15-241121 and czech-pdt-ud-2.12-230717), a widely used, strong baseline that performed at or near the top in relevant shared tasks and supports the UD ecosystem end to end (\cite{straka2018udpipe2}, \url{https://ufal.mff.cuni.cz/udpipe/2}). Output is serialized to standard \emph{CoNLL-U} \citep{ud_conllu_format} accompanied with model-related metadata (e.g., provider/model IDs, sampling temperature, system prompt, generation time, token log-probability if available) or metadata related to the original text that served as a prompt (author, title, text type, publication year).

For corpus-management interoperability (including integration with the Czech National Corpus infrastructure), we additionally export a verticalized representation, a simple flat format that is easy to parse and index. This representation is natively supported by \emph{Manatee} and \emph{IMS Open Corpus Workbench (CWB)} backends and their web interfaces (\emph{KonText}, \emph{CQPweb}, and Sketch Engine family) \citep{rychly2004manatee,rychly2007manatee,evert2011cwb,machalek2020kontext}.

\section{Results}\label{sec:results}

\subsection{Produced text chunks}

The typical failure mode of autoregressive language models (not only Transformer-based ones) is that with longer outputs they either \emph{drift} outside the initial prompt topic, style, or even language, or \emph{loop} locally (repeating the same sentence or paragraph). In our corpus, at sampling temperature $T{=}1$ models tended to drift, whereas with deterministic temperature $T{=}0$ we occasionally observed mode collapse of the latter type. In English, nearly all model–temperature combinations produced at least partially coherent continuations (with the exception of the oldest \texttt{davinci-002}, which tended to loop at the sentence level).

In Czech, weaker models were largely unusable and the above issues affected even stronger systems. With $T{=}0$, \texttt{davinci-002} looped in Czech just as it did in English; similar behavior was observed for the \emph{completion mode} of \texttt{GPT-3.5-turbo}. Czech appears to have been underrepresented in the training data of \textit{Llama 3.1}: neither the base nor the instruction-tuned variant yielded usable continuations. Accordingly, these models were either not used for generation at all or excluded from the distributed corpus (but their outputs are available for download at the same locations as the accepted texts, in a dedicated \emph{Failed} folder --- this applies, e.g., to \textit{Llama 3.1 Instruct} and \textit{GPT-3.5-turbo}).

At $T{=}1$ in Czech, \texttt{davinci-002} often produced wild, semantically incoherent and sometimes agrammatical sequences reminiscent of nonsense poetry. We nevertheless included these outputs because these texts constitute an interesting historical artifact.

\begin{figure}
\centering
\includegraphics[width=0.9\linewidth]{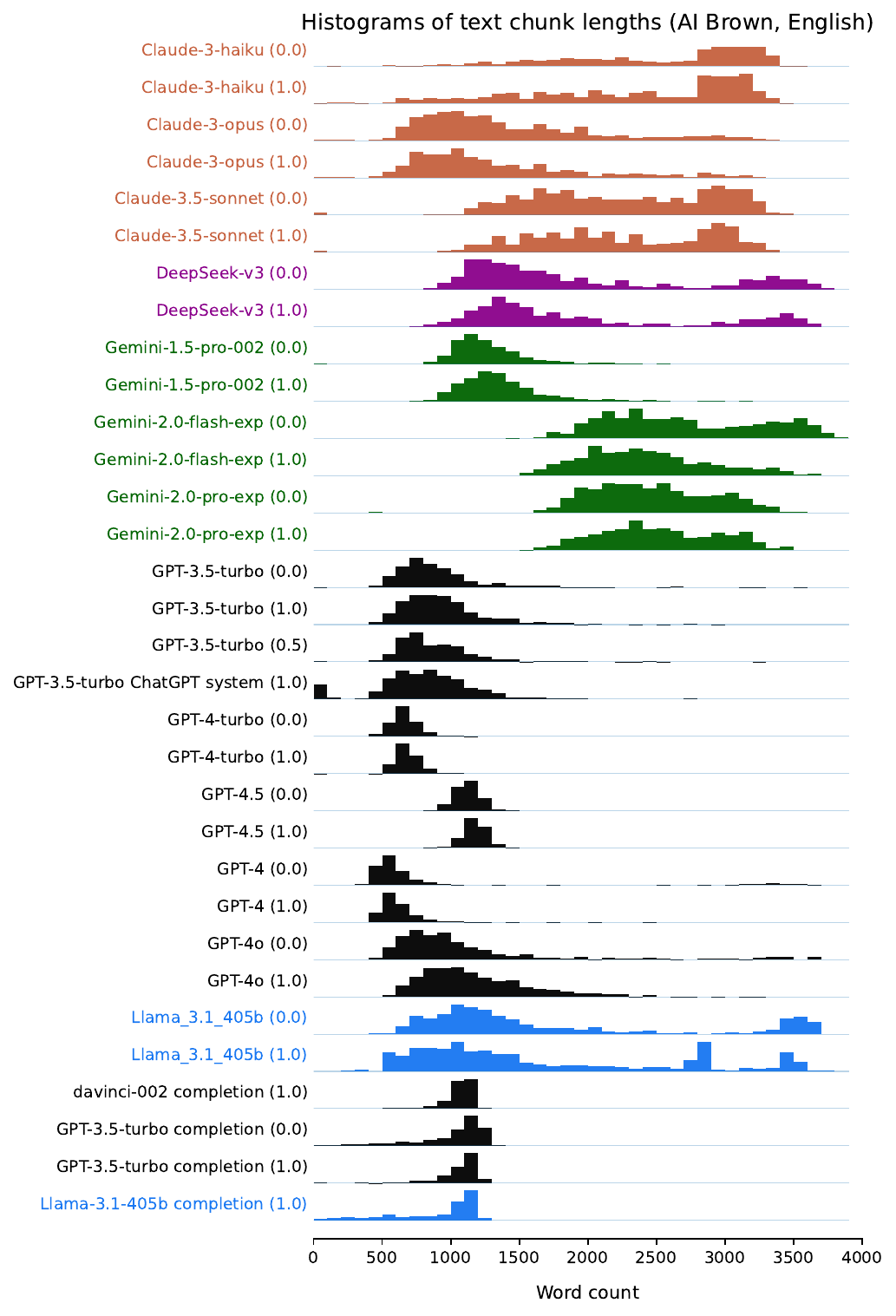}
\caption{AI Brown: Distribution of the generated text chunk lenghts.}
\label{fig:distBrown}
\end{figure}

\begin{figure}
\centering
\includegraphics[width=0.99\linewidth]{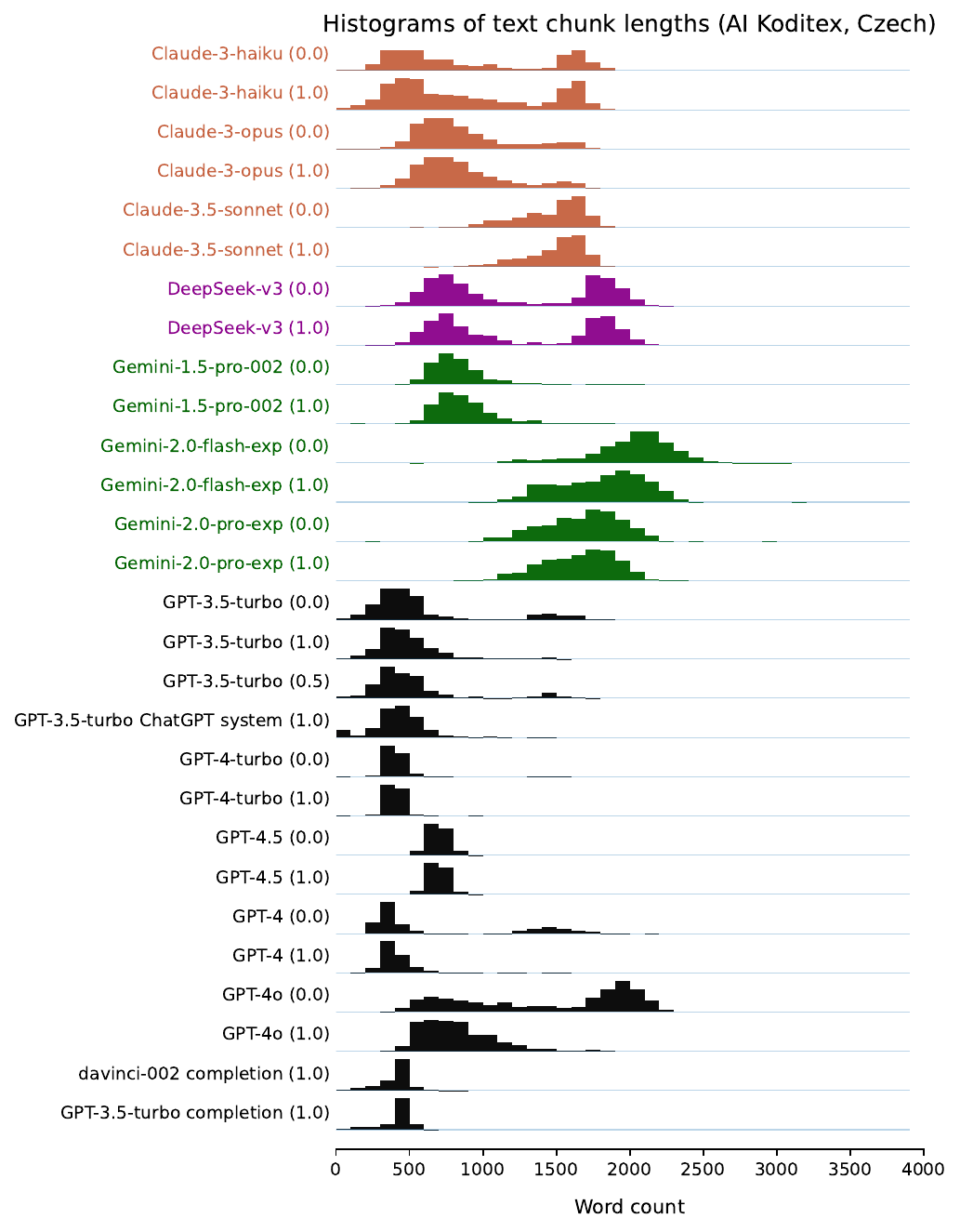}
\caption{AI Koditex: Distribution of the generated text chunk lenghts.}
\label{fig:distKoditex}
\end{figure}

As upper lenght limits we set 1{,}400 tokens for older models and 4{,}096 tokens for newer ones. In our data this corresponds to roughly 3{,}500 word tokens for English but a much shorter limit for Czech (around 2{,}500 word tokens), since subword tokenization is Anglocentric and, on average, consumes more tokens per Czech word. In many cases, however, models terminated before reaching the ceiling, yielding a frequently multi-modal distribution of word-token counts, as shown in Figures~\ref{fig:distBrown} and \ref{fig:distKoditex}.

For the subcorpora generated with the original long ChatGPT system prompt, we observe a nontrivial mass of samples under 100 words. This reflects cases where the model refused to continue and produced a creative justification instead. The frequency of refusals is best seen in Tables~\ref{tab:Brown_model-summary} and \ref{tab:Koditex_model-summary} (column \textit{Refusal}).

These tables also list other key attributes: exact model identifier, generation dates, and total subcorpus size. Length in \emph{words} is computed excluding punctuation and tokens that UDPipe assigns POS ``\_'' or ``X''; length in \emph{positions} is the total number of tokens segmented by the UDPipe tokenizer. Since overall corpus size depends on tokenization method, other tools may report slightly different values.

\afterpage{%
  \clearpage
  \newgeometry{margin=1.5cm}
  \begin{landscape}
  \small
\begin{table}[htbp]
\centering
\small
\begin{tabular}{
  l
  l
  S[table-format=1.1]
  l
  l
  S[table-format=7.0]
  S[table-format=7.0]
  S[table-format=2.1]
}
\toprule
\makebox[0pt][l]{\textbf{ID}} & \textbf{Label} & {\textbf{Temp.}} & \textbf{Vendor} & \textbf{Gen. date} & {\textbf{Words}} & {\textbf{Positions}} & {\textbf{Refusal (\%)}} \\
\midrule
\texttt{brown\_complete\_davinci-002\_\_100} & davinci-002 completion & 1.0 & OpenAI & 29.6.2024 & 532938 & 624417 & 0.0 \\
\texttt{brown\_complete\_llama\_31\_405b\_\_100} & Llama-3.1-405b completion & 1.0 & Meta & 29.12.2024 & 451314 & 526982 & 1.6 \\
\texttt{brown\_complete\_gpt-35-turbo\_\_0} & GPT-3.5-turbo completion & 0.0 & OpenAI & 29.6.2024 & 505158 & 602196 & 0.4 \\
\texttt{brown\_complete\_gpt-35-turbo\_\_100} & GPT-3.5-turbo completion & 1.0 & OpenAI & 2.7.2024 & 527157 & 617374 & 0.4 \\
\texttt{brown\_chat\_gpt-35-turbo\_\_0} & GPT-3.5-turbo & 0.0 & OpenAI & 29.6.2024 & 451527 & 517739 & 0.2 \\
\texttt{brown\_chat\_gpt-35-turbo\_\_100} & GPT-3.5-turbo & 1.0 & OpenAI & 30.6.2024 & 478886 & 549022 & 0.0 \\
\texttt{brown\_chat\_gpt-35-turbo\_\_50} & GPT-3.5-turbo & 0.5 & OpenAI & 30.6.2024 & 446184 & 511786 & 0.4 \\
\texttt{brown\_chat\_gpt-35-turbo\_default-system\_100} & GPT-3.5-turbo ChatGPT system & 1.0 & OpenAI & 1.7.2024 & 408086 & 468697 & 7.0 \\
\texttt{brown\_chat\_gpt-4\_\_0} & GPT-4 & 0.0 & OpenAI & 1.7.2024 & 417764 & 489468 & 0.0 \\
\texttt{brown\_chat\_gpt-4\_\_100} & GPT-4 & 1.0 & OpenAI & 1.7.2024 & 313447 & 364756 & 0.0 \\
\texttt{brown\_chat\_gpt-4-turbo\_\_0} & GPT-4-turbo & 0.0 & OpenAI & 30.6.2024 & 330863 & 382493 & 0.0 \\
\texttt{brown\_chat\_gpt-4-turbo\_\_100} & GPT-4-turbo & 1.0 & OpenAI & 30.6.2024 & 341833 & 394518 & 0.2 \\
\texttt{brown\_chat\_gpt-4o\_\_0} & GPT-4o & 0.0 & OpenAI & 30.6.2024 & 579277 & 677895 & 0.0 \\
\texttt{brown\_chat\_gpt-4o\_\_100} & GPT-4o & 1.0 & OpenAI & 30.6.2024 & 600341 & 704233 & 0.0 \\
\texttt{brown\_chat\_gpt-45\_\_0} & GPT-4.5 & 0.0 & OpenAI & 23.5.2025 & 559073 & 661957 & 0.0 \\
\texttt{brown\_chat\_gpt-45\_\_100} & GPT-4.5 & 1.0 & OpenAI & 23.5.2025 & 586688 & 688280 & 0.0 \\
\texttt{brown\_chat\_claude-3-haiku\_\_0} & Claude-3-haiku & 0.0 & Anthropic & 29.6.2024 & 1286575 & 1496285 & 0.4 \\
\texttt{brown\_chat\_claude-3-haiku\_\_100} & Claude-3-haiku & 1.0 & Anthropic & 30.6.2024 & 1152307 & 1344496 & 0.4 \\
\texttt{brown\_chat\_claude-3-opus\_\_0} & Claude-3-opus & 0.0 & Anthropic & 2.7.2024 & 704734 & 818522 & 0.8 \\
\texttt{brown\_chat\_claude-3-opus\_\_100} & Claude-3-opus & 1.0 & Anthropic & 2.7.2024 & 664986 & 770822 & 0.8 \\
\texttt{brown\_chat\_claude-35-sonnet\_\_0} & Claude-3.5-sonnet & 0.0 & Anthropic & 30.6.2024 & 1132537 & 1331406 & 0.6 \\
\texttt{brown\_chat\_claude-35-sonnet\_\_100} & Claude-3.5-sonnet & 1.0 & Anthropic & 2.7.2024 & 1120057 & 1318588 & 0.6 \\
\texttt{brown\_chat\_deep-seek-v3\_\_0} & DeepSeek-v3 & 0.0 & DeepSeek & 27.12.2024 & 976449 & 1147034 & 0.0 \\
\texttt{brown\_chat\_deep-seek-v3\_\_100} & DeepSeek-v3 & 1.0 & DeepSeek & 27.12.2024 & 980510 & 1153420 & 0.0 \\
\texttt{brown\_chat\_llama\_31\_405b\_\_0} & Llama\_3.1\_405b & 0.0 & Meta & 29.12.2024 & 850037 & 984856 & 0.0 \\
\texttt{brown\_chat\_llama\_31\_405b\_\_100} & Llama\_3.1\_405b & 1.0 & Meta & 3.1.2025 & 828073 & 960888 & 0.0 \\
\texttt{brown\_chat\_gemini-15-pro-002\_\_0} & Gemini-1.5-pro-002 & 0.0 & Alphabet & 29.12.2024 & 633508 & 747093 & 0.4 \\
\texttt{brown\_chat\_gemini-15-pro-002\_\_100} & Gemini-1.5-pro-002 & 1.0 & Alphabet & 30.12.2024 & 686719 & 809877 & 0.0 \\
\texttt{brown\_chat\_gemini-20-flash-exp\_\_0} & Gemini-2.0-flash-exp & 0.0 & Alphabet & 30.12.2024 & 1346948 & 1569273 & 0.6 \\
\texttt{brown\_chat\_gemini-20-flash-exp\_\_100} & Gemini-2.0-flash-exp & 1.0 & Alphabet & 30.12.2024 & 1207795 & 1414078 & 0.2 \\
\texttt{brown\_chat\_gemini-20-pro-exp\_\_0} & Gemini-2.0-pro-exp & 0.0 & Alphabet & 7.2.2025 & 1237059 & 1492789 & 0.2 \\
\texttt{brown\_chat\_gemini-20-pro-exp\_\_100} & Gemini-2.0-pro-exp & 1.0 & Alphabet & 7.2.2025 & 1255607 & 1520202 & 0.2 \\
\bottomrule
\end{tabular}
\caption{Brown subcorpora summary with temperatures, vendors, dates, token counts, and refusal rates.}
\label{tab:Brown_model-summary}
\end{table}
  \end{landscape}
    \restoregeometry
  \clearpage
}

\afterpage{%
  \clearpage
  \newgeometry{margin=1.5cm}
  \begin{landscape}
  \small
\begin{table}[htbp]
\centering
\small
\begin{tabular}{
  l
  l
  S[table-format=1.1]
  l
  l
  S[table-format=7.0]
  S[table-format=7.0]
  S[table-format=2.1]
}
\toprule
\makebox[0pt][l]{\textbf{ID}} & \textbf{Label} & {\textbf{Temp.}} & \textbf{Vendor} & \textbf{Gen. date} & {\textbf{Words}} & {\textbf{Positions}} & {\textbf{Refusal (\%)}} \\
\midrule
\texttt{koditex\_complete\_davinci-002\_\_100} & davinci-002 completion & 1.0 & OpenAI & 2.7.2024 & 272848 & 428186 & 0.7 \\
\texttt{koditex\_complete\_gpt-35-turbo\_\_100} & GPT-3.5-turbo completion & 1.0 & OpenAI & 3.7.2024 & 283499 & 414678 & 1.2 \\
\texttt{koditex\_chat\_gpt-35-turbo\_\_0} & GPT-3.5-turbo & 0.0 & OpenAI & 30.6.2024 & 397062 & 467287 & 1.3 \\
\texttt{koditex\_chat\_gpt-35-turbo\_\_100} & GPT-3.5-turbo & 1.0 & OpenAI & 30.6.2024 & 327200 & 382932 & 0.9 \\
\texttt{koditex\_chat\_gpt-35-turbo\_\_50} & GPT-3.5-turbo & 0.5 & OpenAI & 1.7.2024 & 354946 & 416782 & 1.3 \\
\texttt{koditex\_chat\_gpt-35-turbo\_default-system\_100} & GPT-3.5-turbo ChatGPT system & 1.0 & OpenAI & 2.7.2024 & 288088 & 343305 & 7.1 \\
\texttt{koditex\_chat\_gpt-4\_\_0} & GPT-4 & 0.0 & OpenAI & 2.7.2024 & 481873 & 583010 & 0.0 \\
\texttt{koditex\_chat\_gpt-4\_\_100} & GPT-4 & 1.0 & OpenAI & 2.7.2024 & 282823 & 336083 & 0.0 \\
\texttt{koditex\_chat\_gpt-4-turbo\_\_0} & GPT-4-turbo & 0.0 & OpenAI & 30.6.2024 & 272442 & 320120 & 0.3 \\
\texttt{koditex\_chat\_gpt-4-turbo\_\_100} & GPT-4-turbo & 1.0 & OpenAI & 1.7.2024 & 267396 & 313578 & 0.6 \\
\texttt{koditex\_chat\_gpt-4o\_\_0} & GPT-4o & 0.0 & OpenAI & 1.5.2025 & 965428 & 1154264 & 0.0 \\
\texttt{koditex\_chat\_gpt-4o\_\_100} & GPT-4o & 1.0 & OpenAI & 1.7.2024 & 557861 & 658671 & 0.0 \\
\texttt{koditex\_chat\_gpt-45\_\_0} & GPT-4.5 & 0.0 & OpenAI & 6.6.2025 & 470160 & 556518 & 0.0 \\
\texttt{koditex\_chat\_gpt-45\_\_100} & GPT-4.5 & 1.0 & OpenAI & 8.6.2025 & 470395 & 549276 & 0.0 \\
\texttt{koditex\_chat\_claude-3-haiku\_\_0} & Claude-3-haiku & 0.0 & Anthropic & 30.6.2024 & 615910 & 713067 & 0.3 \\
\texttt{koditex\_chat\_claude-3-haiku\_\_100} & Claude-3-haiku & 1.0 & Anthropic & 30.6.2024 & 595001 & 686046 & 0.9 \\
\texttt{koditex\_chat\_claude-3-opus\_\_0} & Claude-3-opus & 0.0 & Anthropic & 2.7.2024 & 585452 & 689553 & 0.3 \\
\texttt{koditex\_chat\_claude-3-opus\_\_100} & Claude-3-opus & 1.0 & Anthropic & 3.7.2024 & 575283 & 677052 & 0.0 \\
\texttt{koditex\_chat\_claude-35-sonnet\_\_0} & Claude-3.5-sonnet & 0.0 & Anthropic & 1.7.2024 & 990959 & 1180664 & 0.0 \\
\texttt{koditex\_chat\_claude-35-sonnet\_\_100} & Claude-3.5-sonnet & 1.0 & Anthropic & 2.7.2024 & 1010068 & 1204554 & 0.0 \\
\texttt{koditex\_chat\_deep-seek-v3\_\_0} & DeepSeek-v3 & 0.0 & DeepSeek & 28.12.2024 & 829479 & 996434 & 0.0 \\
\texttt{koditex\_chat\_deep-seek-v3\_\_100} & DeepSeek-v3 & 1.0 & DeepSeek & 28.12.2024 & 836075 & 1005950 & 0.0 \\
\texttt{koditex\_chat\_gemini-15-pro-002\_\_0} & Gemini-1.5-pro-002 & 0.0 & Alphabet & 30.12.2024 & 558581 & 686270 & 0.0 \\
\texttt{koditex\_chat\_gemini-15-pro-002\_\_100} & Gemini-1.5-pro-002 & 1.0 & Alphabet & 30.12.2024 & 583553 & 709096 & 0.0 \\
\texttt{koditex\_chat\_gemini-20-flash-exp\_\_0} & Gemini-2.0-flash-exp & 0.0 & Alphabet & 30.12.2024 & 1364057 & 1665263 & 0.0 \\
\texttt{koditex\_chat\_gemini-20-flash-exp\_\_100} & Gemini-2.0-flash-exp & 1.0 & Alphabet & 31.12.2024 & 1214631 & 1464998 & 0.0 \\
\texttt{koditex\_chat\_gemini-20-pro-exp\_\_0} & Gemini-2.0-pro-exp & 0.0 & Alphabet & 7.2.2025 & 1131666 & 1467355 & 0.4 \\
\texttt{koditex\_chat\_gemini-20-pro-exp\_\_100} & Gemini-2.0-pro-exp & 1.0 & Alphabet & 7.2.2025 & 1112318 & 1430720 & 0.3 \\
\bottomrule
\end{tabular}
\caption{Koditex subcorpora summary with temperatures, vendors, dates, token counts, and refusal rates.}
\label{tab:Koditex_model-summary}
\end{table}
  \end{landscape}
    \restoregeometry
  \clearpage
}

\subsection{Availability of the corpora}

Crucially, the corpora are downloadable from the LINDAT/CLARIAH-CZ linguistic repository, which provides structured metadata and persistent identifiers. The raw data and cleaned plain texts are published under CC BY-4.0 licence, while the UDPipe processed files (namely CoNLL-U files and verticals) are under CC BY-NC-SA licence (the requirement of UDPipe).

Within this repository the corpora are available in the following formats (all UTF-8 encoded):

\paragraph{Raw outputs of models.} Located in \texttt{raw.zip}; these JSON files include, wherever possible, all information obtainable from the API interaction. For older OpenAI endpoints this includes alternative tokens and their log-probabilities.

\paragraph{Plain texts.} Extracted from the raw JSON and stored in \texttt{plain\_texts.zip}. In addition to the generated text, these files may contain meta-prefaces that instruction-tuned models prepend and/or refusal rationales (see Section~\ref{sec:postprocessing}).

\paragraph{Cleaned texts.} Stored in \texttt{plain\_texts\_cleaned.zip}. These are the most convenient starting point for downstream work: meta-prefaces and refusals have been stripped, but no linguistic tags are added.

\paragraph{UD-tagged texts.} The tokenized, lemmatized, morphologically and syntactically annotated texts are provided in CoNLL-U format and stored in \texttt{UD.zip}.

\paragraph{UD-tagged texts in vertical representation.} Stored in \texttt{vertical.zip}. Since some systems still do not natively process CoNLL-U, we also provide a verticalized representation, the long-standing de facto standard in corpus linguistics.

Additionally, the corpora are searchable on the infrastructure of the Czech National Corpus via the KonText interface at \url{https://www.korpus.cz/kontext/query?corpname=ai_brown_v1} and \url{https://www.korpus.cz/kontext/query?corpname=ai_koditex_v1}. Since annotation follows Universal Dependencies, users can query not only by lemmas and morphological features but also by syntactic relations via the Corpus Query Language (CQL). As the CQL can be demanding, we recommend the Alpha tool \url{http://alpha.korpus.cz/} ( see detailes in \cite{milivcka2024chapter}) for translating natural-language queries into CQL. For example, the request \emph{find all nouns that are dependent on lemma ``delve''} is translated into \texttt{[upos="NOUN|PROPN" \& p\_lemma="delve"]}. When entered under \emph{advanced query} in KonText, this retrieves a variety of concepts that models typically ``delve into''. KonText can also calculate basic distributional statistics, therefore the user can then inspect which such delved-into items are most frequent.

\subsection{Case studies}
\subsubsection{Formulaicity}

We now illustrate various kinds of analyses these corpora enable.

One question concerns the \emph{formulaicity} of LLM language. Consider a prominent phrase familiar to many users of ChatGPT and ChatGPT-like interfaces: \emph{testament to}. In the AI-Brown corpus, the bigram \emph{testament to} is extremely frequent (222 instances per million, i.p.m.), and --- crucially --- the longer string \emph{a testament to the} occurs very often as well (163 i.p.m.). By contrast, the exact string \emph{a testament to the} does not occur at all in the original BE21 reference corpus, and the shorter \emph{testament to} appears only 7 i.p.m.

\begin{figure}
    \centering
    \includegraphics[width=0.99\linewidth]{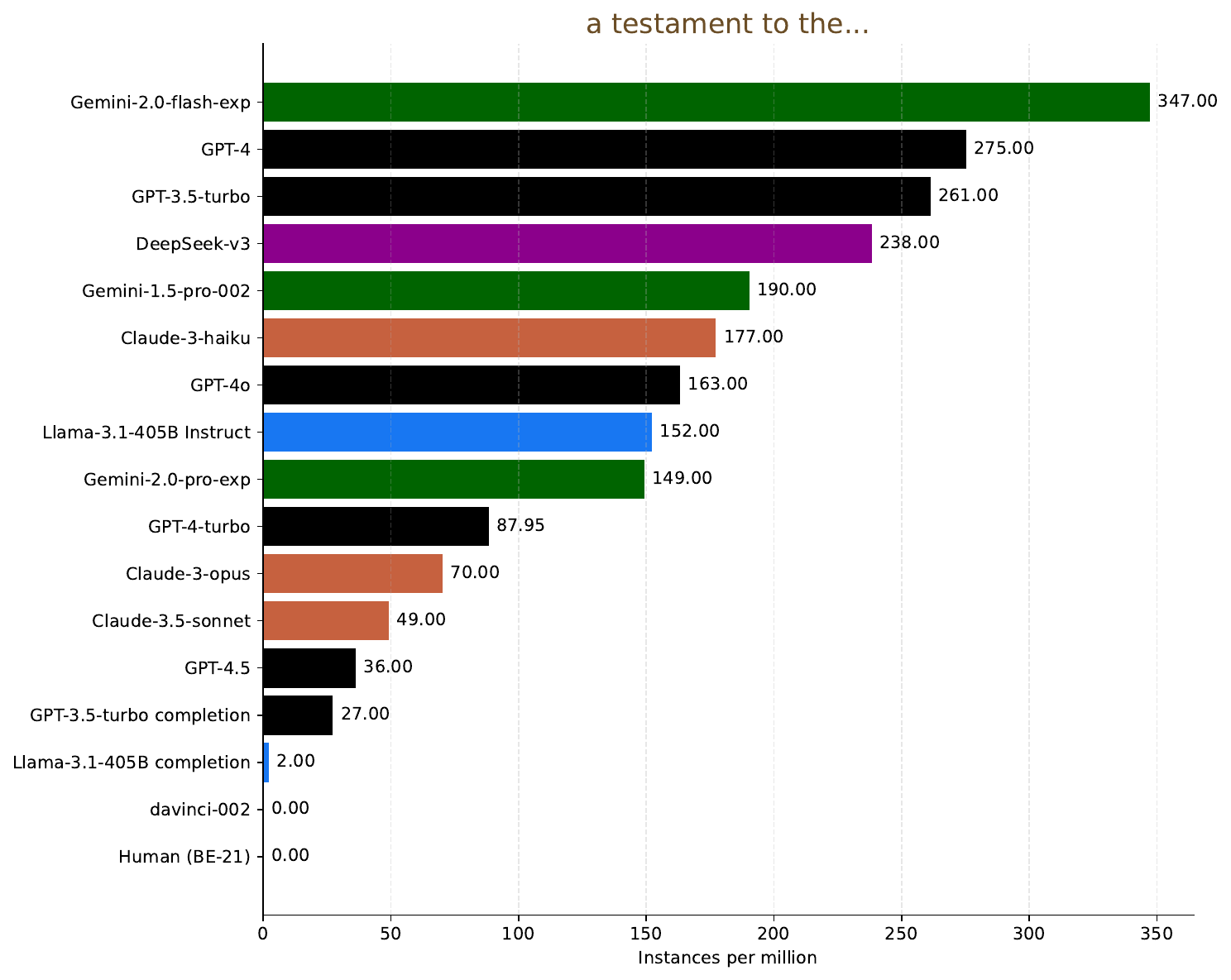}
    \caption{Instances per million of \emph{a testament to the} in subcorpora of AI Brown and in the original human BE21 corpus.}
    \label{fig:ATestamentToTheBrown}
\end{figure}

Examining subcorpora shows that this formulaic expression emerges primarily with \emph{instruction-tuned} systems; it is rare or absent in base models. Strikingly, the phrase appears across instruction-tuned variants from different companies. A plausible explanation is that instruction tuning often leverages synthetic data distilled from other models, allowing such formulas to propagate from model to model (see Figure~\ref{fig:ATestamentToTheBrown}).

Formulaicity is even more pronounced in AI Czech. This is partly because several models --- especially the Gemini variants --- struggled with Czech and occasionally fell into repetitive loops. We examine this in more detail below.

In the human-authored Koditex subcorpus, 4-grams are relatively rare; the ten most frequent sequences are listed in Table~\ref{tab:pn-4grams-no-model}.

\afterpage{%
  \clearpage
  \newgeometry{margin=1.5cm}
  \begin{landscape}
\begin{table}[htbp]
\centering
\small
\begin{tabular}{
  l
  l
  S[table-format=5.0]
  l
  S[table-format=3.2]
  l
}
\toprule
\textbf{4-gram} & \textbf{English translation} & {\textbf{Freq}} & \textbf{CI (freq)} & {\textbf{i.p.m.}} & \textbf{CI (i.p.m.)} \\
\midrule
bez ohledu na to & regardless & 160 & [137, 187] & 14.71 & [13, 17] \\
já si myslím že & I think that & 140 & [119, 165] & 12.87 & [11, 15] \\
já si myslim že & I think that & 133 & [112, 158] & 12.22 & [10, 14] \\
že se jedná o & that it is about & 113 & [94, 136] & 10.39 & [9, 12] \\
ve znění pozdějších předpisů & as amended & 105 & [87, 127] & 9.65 & [8, 12] \\
a v neposlední řadě & and last but not least & 97 & [80, 118] & 8.92 & [7, 11] \\
a od té doby & and since then & 93 & [76, 114] & 8.55 & [7, 10] \\
ale na druhou stranu & but on the other hand & 84 & [68, 104] & 7.72 & [6, 10] \\
to je to je & that's that's & 81 & [65, 101] & 7.44 & [6, 9] \\
od té doby se & since then & 81 & [65, 101] & 7.44 & [6, 9] \\

\bottomrule
\end{tabular}
\caption{Top Czech 4-grams (with English translations) from the original human Koditex, absolute frequencies, instances per million, and 95 \% confidence intervals.}
\label{tab:pn-4grams-no-model}
\end{table}

\begin{table}[htbp]
\centering
\small
\begin{tabular}{
  l
  l
  S[table-format=5.0]
  l
  S[table-format=3.2]
  l
}
\toprule
\textbf{4-gram}& \textbf{English translation} & {\textbf{Freq}} & \textbf{CI (freq)} & {\textbf{i.p.m.}} & \textbf{CI (i.p.m.)} \\
\midrule

a to je to & and that is the & 7393 & [7226, 7563] & 345.87 & [338, 354] \\
a co se týče & and as for & 2683 & [2583, 2786] & 125.52 & [121, 130] \\
ať už jde o & whether it is about & 1527 & [1452, 1606] & 71.44 & [68, 75] \\
je důležité si uvědomit & it is important to realize & 1523 & [1448, 1601] & 71.25 & [68, 75] \\
a to je něco & and that is something & 1457 & [1384, 1534] & 68.16 & [65, 72] \\
v neposlední řadě je & last but not least is & 1307 & [1238, 1380] & 61.15 & [58, 65] \\
proti proti proti proti & against against against against & 1017 & [956, 1081] & 47.58 & [45, 51] \\
jaký je váš názor & what is your opinion & 994 & [934, 1058] & 46.50 & [44, 49] \\
je váš názor na & is your opinion on & 990 & [930, 1054] & 46.32 & [44, 49] \\
už se jedná o & wheather it is about & 889 & [832, 949] & 41.59 & [39, 44] \\

\bottomrule
\end{tabular}
\caption{Top Czech 4-grams (with English translations) from AI Koditex, absolute frequencies, instances per million, and 95 \% confidence intervals.}
\label{tab:pn-4grams-top10}
\end{table}

  \end{landscape}
    \restoregeometry
  \clearpage
}

As Table~\ref{tab:pn-4grams-no-model} shows, the most frequent 4-gram, \emph{bez ohledu na to} ``regardless'', occurs at 14.71 instances per million (i.p.m.). The second and third entries are \emph{já si myslím že} ``I think that'' and its colloquial variant \emph{já si myslim že}, which differ only in vowel length and essentially realize the same construction. Taken together, they reach 25.09 i.p.m. (12.87 + 12.22), still far below what we observe in the AI Koditex corpus (Table~\ref{tab:pn-4grams-top10}).

In AI Koditex, formulaicity is markedly higher than in Koditex. The top 4-gram \emph{a to je to} ``and that is the'' reaches 346 i.p.m., followed by \emph{a co se týče} ``as for/with regard to'' at 126 i.p.m. A close inspection indicates that these peaks are largely driven by repetition in Gemini outputs: within \texttt{gemini-2.0-flash-exp}, \emph{a to je to} ``and that is the'' reaches 1{,}921 i.p.m., compared with 285 i.p.m. in \texttt{deepseek-chat}. Likewise, \emph{a co se týče} ``as for/with regard to'' has 653 i.p.m. in \texttt{gemini-2.0-flash-exp}, 101 i.p.m. in \texttt{gemini-2.0-pro-exp}, and 67 i.p.m. in \texttt{deepseek-chat}. The anomalous sequence \emph{proti proti proti proti} ``against against against against'' stems from a collapse in \texttt{gemini-1.5-pro-002} while generating a particular text.

Outside these Gemini-driven outliers, the remaining high-frequency 4-grams are distributed more evenly across models. For example, \emph{ať už jde o} ``whether it is about'' is highest in \texttt{gpt-4-turbo} (178 i.p.m.), followed by \texttt{deepseek-chat} (144 i.p.m.), then \texttt{claude-3-haiku} and \texttt{claude-3-5-sonnet-20240620} (113 i.p.m.), etc.

Overall, \texttt{gemini-2.0-flash-exp}, \texttt{gemini-2.0-pro-exp-02-05}, and \texttt{deepseek-chat} consistently rank near the top in formulaic n-gram usage, whereas models such as \texttt{gpt-4.5} and \texttt{gpt-3.5-turbo-instruct} rank much lower. Table~\ref{tab:czech-4grams} summarizes the most frequent 4-grams by model.

\afterpage{%
  \newgeometry{margin=1.5cm}
  \begin{landscape}

\begin{table}[htbp]
\centering
\small
\begin{tabular}{
  l
  l
  l
  S[table-format=5.0]
  l
  S[table-format=3.2]
  l
}
\toprule
\textbf{Model} & \textbf{4-gram}& \textbf{English translation} & {\textbf{Freq}} & \textbf{CI (freq)} & {\textbf{i.p.m.}} & \textbf{CI (i.p.m.)} \\
\midrule
gemini-2.0-flash-exp & a to je to & and that is the & 5966 & [5817, 6119] & 279.11 & [272, 286] \\
deepseek-chat & a to je to & and that is the & 565 & [520, 614] & 26.43 & [24, 29] \\
gemini-2.0-pro-exp & a to je to & and that is the & 437 & [398, 480] & 20.44 & [19, 22] \\
gpt-4 & je váš názor na & is your opinion on & 431 & [392, 474] & 20.16 & [18, 22] \\
gpt-4o & 24 pokračování ze str & 24 continuation from p. & 291 & [259, 326] & 13.61 & [12, 15] \\
claude-3-5-sonnet & ať už jde o & whether it is about & 269 & [239, 303] & 12.59 & [11, 14] \\
claude-3-haiku & už se jedná o & it is already about & 200 & [174, 230] & 9.36 & [8, 11] \\
gpt-3.5-turbo & ať už je to & whether it is & 195 & [169, 224] & 9.12 & [8, 11] \\
gemini-1.5-pro-002 & v neposlední řadě je & last but not least is & 171 & [147, 199] & 8 & [7, 9] \\
gpt-4-turbo & ať už jde o & whether it is about & 112 & [93, 135] & 5.24 & [4, 6] \\
claude-3-opus & ať už jde o & whether it is about & 110 & [91, 133] & 5.15 & [4, 6] \\
davinci-002 & když i jednou míni & when even once means & 98 & [80, 119] & 4.59 & [4, 6] \\
gpt-3.5-turbo-instruct & neplodným se české republiky & unfruitful of the Czech Republic & 80 & [64, 100] & 3.74 & [3, 5] \\
gpt-4.5 & ať už jde o & whether it is about & 66 & [52, 84] & 3.09 & [2, 4] \\

\bottomrule
\end{tabular}
\caption{Most frequent Czech 4-grams by model, with English translations, absolute frequencies, instances per million, and 95 \% confidence intervals.}
\label{tab:czech-4grams}
\end{table}
  \end{landscape}
    \restoregeometry
  \clearpage
}





\subsubsection{Stylistic variability}

An important question about the language of large language models is whether they can produce \emph{stylistically diverse} texts. Since the human source corpora used to seed AI continuations are themselves diverse and genre-annotated, we can test how faithfully AI continuations adhere to the styles of their respective genres. In \citep{milika2025benchmark}, we conducted a Biber-style multidimensional analysis and established a stylistic benchmark for evaluating AI texts. We found that: (1) base models and instruction-tuned models differ, with base models generally performing better than instruction-tuned ones; (2) performance is better in English than in Czech, likely because English dominates the training data; and (3) while all models exhibit distinct overall shifts and genre-specific shifts (i.e., there is no single ``AI language''), some stylistic dimensions are generally easier for models to mimic than others (e.g., narrativity appears easier to approximate than non-explicit reference).

Model-specific shifts across temperatures, prompts, and versions are available as interactive visualizations in the online interface: \url{https://korpus.cz/stylisticbenchmark/}

\subsubsection{Effect of sampling temperature on lexical diversity}

Sampling temperature should affect lexical diversity: higher temperatures are expected to yield more variable outputs and thus richer vocabularies --- a pattern repeatedly observed in prior work \citep{martinez2024beware,guo2024benchmarking}. We can test this hypothesis directly with our corpus. We operationalize lexical diversity as the \emph{moving-average type–token ratio} (MATTR, see \cite{covington2010cutting}), which calculates lexical richness over a sliding window. For our purposes, we use a window size of 100 word tokens to accommodate shorter texts, though larger windows would also be feasible.

\begin{figure}
    \centering
    \includegraphics[width=0.99\linewidth]{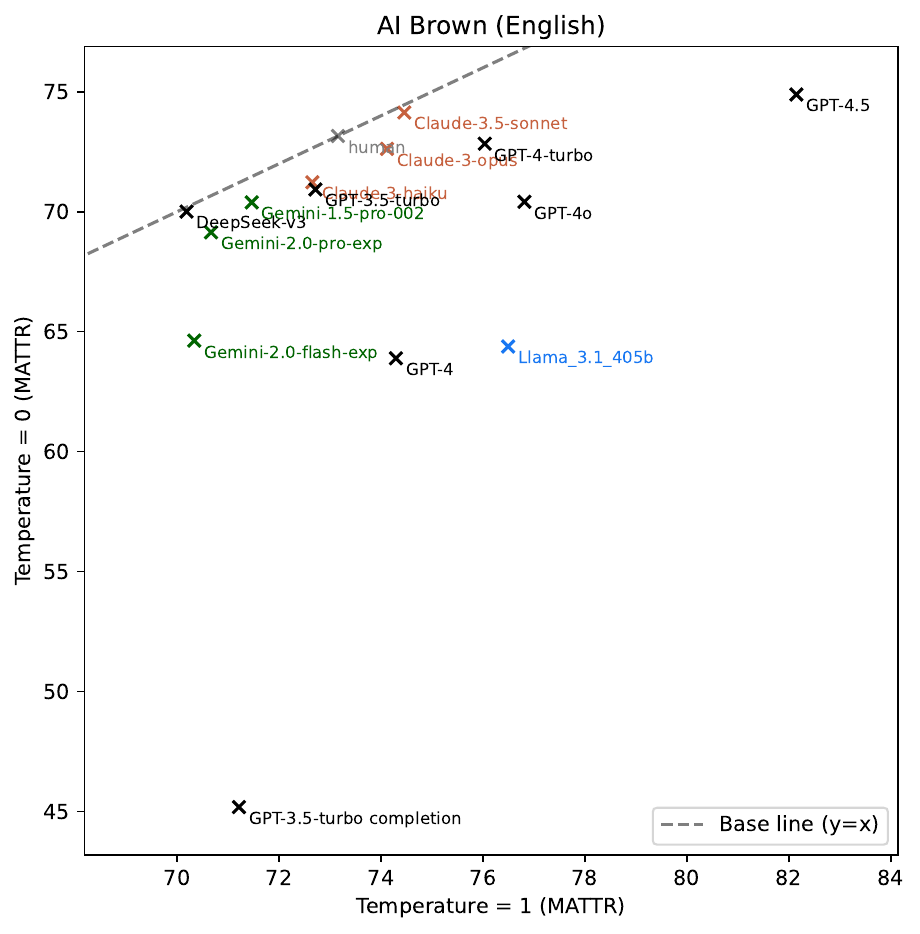}
    \caption{AI Brown: Each dot represents one model, on x axis is its MATTR when the sampling temperature is 1, on y axis is its MATTR when the sampling temperature is 0.}
    \label{fig:scatterBrown}
\end{figure}

\begin{figure}
    \centering
    \includegraphics[width=0.99\linewidth]{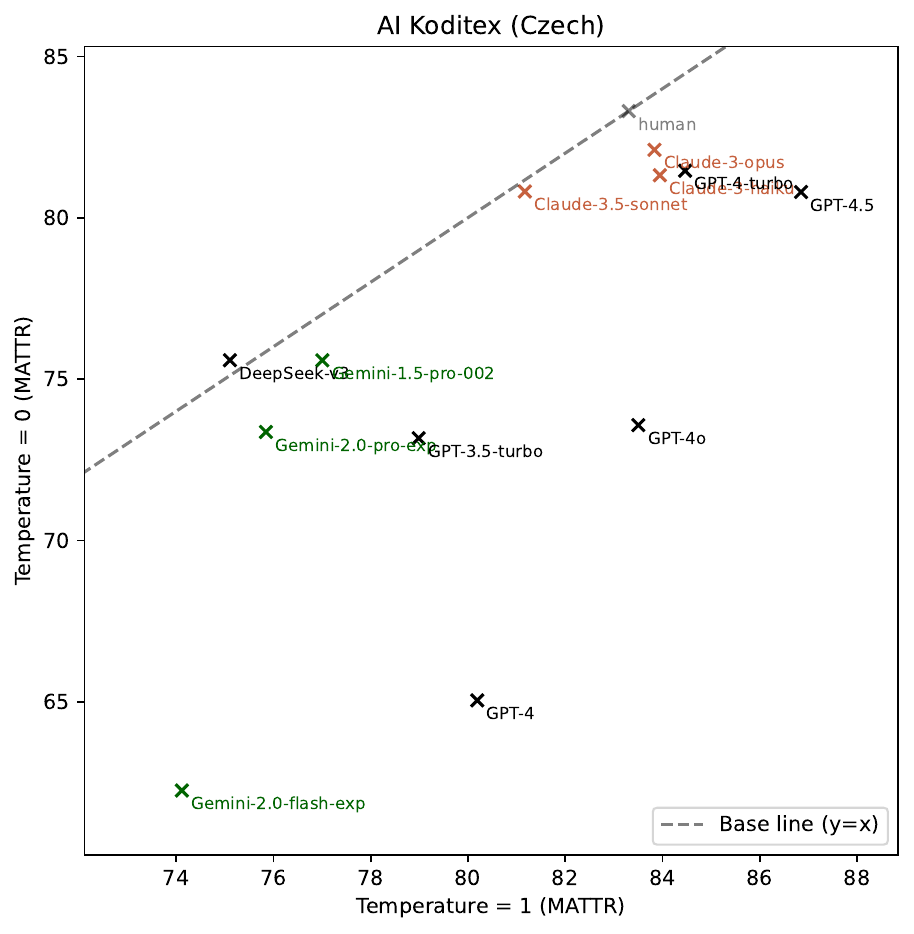}
    \caption{AI Koditex: Each dot represents one model, on x axis is its MATTR when the sampling temperature is 1, on y axis is its MATTR when the sampling temperature is 0.}
    \label{fig:scatterCzech}
\end{figure}

The results, shown in Figure~\ref{fig:scatterBrown} for English and Figure~\ref{fig:scatterCzech} for Czech, indicate that, in all cases, lexical diversity at $T{=}1$ is greater than or equal to that at $T{=}0$ (with the sole exception of DeepSeek in Czech). We also observe that some models exhibit lexical diversity close to that of human texts under both temperatures (notably several Anthropic models).

\section{Conclusion}

The primary goal of these corpora is to lower the barrier to research on AI-generated language for scholars whose expertise lies in other areas (morphology, syntax, lexicography, discourse, etc.) and who wish to test hypotheses without investing time in assembling ad hoc datasets. We also hope the corpora will prove useful at forthcoming corpus-linguistics venues that plan to feature LLMs as a central or significant theme (e.g., forthcoming ICAME 2026 in Koblenz).

Thanks to a permissive license, users are invited to fork the corpora and add further layers of annotation. We also release the scripts used to generate the datasets, making it straightforward to create comparable corpora --- either in other languages or with more specialized reference corpora.

Finally, we intend to expand these corpora regularly to capture newly emerging and disappearing models over time, thereby serving as a museum of large language models.

\backmatter

\bmhead{Supplementary information}

If your article has accompanying supplementary file/s please state so here. 

Authors reporting data from electrophoretic gels and blots should supply the full unprocessed scans for key as part of their Supplementary information. This may be requested by the editorial team/s if it is missing.

Please refer to Journal-level guidance for any specific requirements.

\bmhead{Acknowledgements}

\section*{Declarations}

\subsection*{Funding}
Jiří Milička was supported by Czech Science Foundation Grant No. 24-11725S, \url{gacr.cz} (``Large language models through the prism of corpus linguistics''). 

Generating of several subcorpora was supported by the project ``Human‐centred AI for a Sustainable and Adaptive Society'' (reg. no.: CZ.02.01.01/00/23\_025/0008691), co‐funded by the European Union.

Anna Marklová was supported by Charles University Grant PRIMUS/25/SSH/010 (``Sensitivity to register variability: A combination of corpus and experimental methodology'').

The implementation of the Corpora in the KonText was supported by the Czech National Corpus project (LM2023044) funded by the Ministry of Education, Youth and Sports of the Czech Republic within the framework of Large Research, Development and Innovation Infrastructures.

The funders had no role in study design, data collection and analysis, decision to publish, or preparation of the manuscript.

\subsection*{Competing interests} 
The authors have no competing interests to declare.

\subsection*{Ethics approval and consent to participate}
`Not applicable'
\subsection*{Consent for publication}
`Not applicable'
\subsection*{Data availability}

AI Brown v1 is available on Lindat: \url{http://hdl.handle.net/11234/1-5993} and also via KonText \url{https://www.korpus.cz/kontext/query?corpname=ai_brown_v1}.

AI Koditex v1 is also available on Lindat: \url{http://hdl.handle.net/11234/1-5991} and via KonText \url{https://www.korpus.cz/kontext/query?corpname=ai_koditex_v1}.

The plain texts are freely available for download under the CC BY 4.0 license, the annotated data are under CC BY-NC-SA 4.0 licence.

\subsection*{Materials availability}
`Not applicable'

\subsection*{Code availability}
The code used to generate and process the corpus and the statistics and visualizations in this paper are available at \url{https://osf.io/sfk6w/}.

\subsection*{Author contribution}
\emph{Jiří Milička}: conceptualization, data curation, funding acquisition, investigation, methodology, project administration, software, resources, supervision, validation, visualization, writing – original draft, writing – review \& editing; \emph{Anna Marklová}: conceptualization, writing – original draft, investigation, writing – review \& editing; \emph{Václav Cvrček}: conceptualization, writing – review \& editing, resources.

\subsection*{Declaration on using AI}
The GPT-4o, GPT-4.5, and GPT-5 models by OpenAI, and Claude 4 Sonnet and Claude 4.1 Opus models by Anthropic were consulted for coding scripts and language editing of the article. All scripts and texts underwent manual review and were corrected or further refined when necessary. The authors assume full responsibility for any errors.

\bibliography{AIcorpora} 

\end{document}